\documentclass[letterpaper,journal]{IEEEtran}
\usepackage{amsmath,amsfonts}
\usepackage{algorithm}
\usepackage{array}
\usepackage[caption=false,font=normalsize,labelfont=sf,textfont=sf]{subfig}
\usepackage{textcomp}
\usepackage{stfloats}
\usepackage{url}
\usepackage{verbatim}
\usepackage{graphicx}
\usepackage{cite}
\usepackage{booktabs}
\usepackage{multirow}
\usepackage{amssymb}
\usepackage{algpseudocode}
\usepackage{xcolor}
\usepackage{pifont}
\DeclareRobustCommand{\NA}{\textcolor{red}{\ding{56}}}

\begin{document}

\title{OTPL-VIO: Robust Visual-Inertial Odometry with Optimal Transport Line Association and Adaptive Uncertainty}

\author{Zikun Chen, Wentao Zhao, Yihe Niu, Tianchen Deng, and Jingchuan Wang,~\IEEEmembership{Senior Member,~IEEE,}

\thanks{Zikun Chen, Wentao Zhao, Tianchen Deng, and Jingchuan Wang are with the School of Automation and Intelligent Sensing, Institute of Medical Robotics, Shanghai Jiao Tong University, Shanghai 200240, China.
Yihe Niu is with the School of Mathematical Sciences Shanghai Jiao Tong University. This work was supported by the National Natural Science Foundation of China under Grant 61773261.
Zikun Chen and Wentao Zhao contributed equally to this work. 
Jingchuan Wang (jchwang@sjtu.edu.cn) is the corresponding author.
}
}

\markboth{IEEE ROBOTICS AND AUTOMATION LETTERS,~Vol.~11, No.~6, June~2026}%
{Shell \MakeLowercase{\textit{et al.}}: A Sample Article Using IEEEtran.cls for IEEE Journals}


\maketitle

\begin{abstract}
Robust stereo visual--inertial odometry (VIO) remains challenging in low-texture scenes and under abrupt illumination changes, where point features become sparse and unstable, leading to ambiguous association and under-constrained estimation. Line structures offer complementary geometric cues, yet many efficient point--line systems still rely on point-guided line association, which can break down when point support is weak and may lead to biased constraints. We present a stereo point--line VIO system in which line segments are equipped with dedicated deep descriptors and matched using an entropy-regularized optimal transport formulation, which performs global mass-transport assignment and supports unmatched observations under ambiguity, outliers, and partial observations. The proposed descriptor is training-free and is computed by sampling and pooling network feature maps. To improve estimation stability, we analyze the impact of line measurement noise and introduce reliability-adaptive weighting to regulate the influence of line constraints during optimization. Experiments on EuRoC and UMA-VI, together with real-world deployments in low-texture and illumination-challenging environments, demonstrate improved accuracy and robustness over representative baselines while maintaining real-time performance.
\end{abstract}


\section{Introduction}
\IEEEPARstart{R}{obust} visual odometry remains difficult in scenes where low texture and abrupt illumination changes co-exist. Low texture reduces the availability of repeatable keypoints, while photometric variations destabilize appearance-based matching. Although deep networks can increase keypoint density in texture-poor regions, these features are often weakly distinctive and frequently fail to satisfy stereo geometric constraints, resulting in triangulation failures. In contrast, line structures are ubiquitous in structured environments and provide complementary geometric constraints that remain informative when point features become sparse or unstable.

Motivated by this, a line of research integrates line segments into SLAM/odometry to enhance robustness, either by exploiting structural priors or by performing joint point--line optimization~\cite{gomez2019pl,jiang2024ul,dong2024stl,chen2024vpl}.
However, many point--line systems still rely on classical line extraction and handcrafted descriptors such as LBD~\cite{zhang2013efficient}, whose discriminability can degrade in low-texture or repetitive regions, leading to fragile data association.
Recent learning-based front-ends improve feature robustness~\cite{wang2017stereo,kannapiran2023stereo,zhao2025uno,zhao2023self}, yet often employ separate heavy networks for different feature types, resulting in redundant computation and hindering real-time deployment.
Moreover, point-guided line association may lose its complementarity when point support becomes weak, while local nearest-neighbor matching can force ambiguous or partially observed line segments into incorrect correspondences. These limitations motivate (i) explicit lightweight line representations, (ii) joint association with unmatched handling, and (iii) reliability-aware optimization for heterogeneous line quality.

To address these issues, this paper presents OTPL-VIO, a robust and efficient stereo point--line visual--inertial odometry system for low-texture and illumination-changing scenes. Different from point-guided or local nearest-neighbor line association, OTPL-VIO equips line segments with an explicit training-free deep descriptor, associates them through entropy-regularized optimal transport with virtual unmatched nodes, and adaptively regulates their back-end influence according to line-specific reliability. The descriptor aggregates features sampled along each detected segment without introducing an additional network; the OT formulation performs joint assignment over candidate segments and allows ambiguous or partially observed segments to remain unmatched; and reliability-adaptive weighting accounts for spatial extent, partial observability, and length-dependent uncertainty. This design aims to make line constraints more reliable when point support becomes weak, while suppressing ambiguous, fragmented, or unreliable line observations.

Overall, our contributions are as follows: 1) A lightweight deep line descriptor that samples and aggregates contextual features along each detected line segment without additional training, providing explicit line representation under low texture and illumination changes; 2) An optimal-transport line association strategy that performs globally normalized joint assignment over candidate segments and supports unmatched segments under ambiguous or partial observations; 3) A reliability-adaptive line weighting strategy that models length-dependent line uncertainty and track persistence, reducing the influence of unreliable line constraints during optimization; 4)  Comprehensive evaluation on EuRoC and UMA-VI benchmarks, as well as real-world deployments in low-texture scenes with abrupt illumination changes, demonstrating improved accuracy and real-time performance.
    
    
    

\section{Related Work}
\subsection{Visual(-Inertial) SLAM/VO under Visual Degradation}
Feature-based SLAM/VO systems estimate motion from sparse keypoints and achieve strong efficiency, with representative pipelines including VINS-Fusion~\cite{qin2018vins}, OKVIS~\cite{leutenegger2015keyframe}, ORB-SLAM3~\cite{campos2021orb}, and Kimera-VIO~\cite{rosinol2020kimera}. 
However, their performance can degrade when visual measurements become unreliable, especially in low-texture regions where repeatable features are scarce.
Direct methods such as DSO~\cite{wang2017stereo} reduce reliance on sparse keypoints by optimizing photometric consistency, but remain sensitive to illumination changes.
Learning-based pipelines such as DROID-SLAM~\cite{teed2021droid} improve robustness in some settings, yet severe visual degradation can still challenge correspondence estimation.
Loop closure and global optimization can reduce long-term drift~\cite{cramariuc2022maplab,peng2024dvi}, but local correspondence failures under visual degradation still motivate complementary geometric primitives such as lines~\cite{gomez2019pl}.

\subsection{Line Features: Detection, Description, and Association}
\noindent\textbf{Detection and description.}
Classical line detectors such as E2LSD~\cite{lin2023effective} are lightweight, but may produce fragmented or less repeatable segments under noise and illumination changes.
Handcrafted descriptors such as LBD~\cite{zhang2013efficient} are widely used for association, but can degrade in low-texture regions with weak gradients.
Learning-based line methods~\cite{pautrat2021sold2,xue2023holistically} improve repeatability under occlusion or viewpoint changes, but their cost may hinder real-time SLAM/VO deployment.

\noindent\textbf{Point--line systems and matching strategies.}
Point--line SLAM/VO systems complement point features with line constraints, including PL-SLAM~\cite{gomez2019pl}, structural or compact variants~\cite{jiang2024ul,chen2024vpl,dong2024stl}, and visual--inertial systems such as PLF-VINS~\cite{lee2021plf} and PL-VINS~\cite{fu2020plvins}.
However, their robustness often depends on reliable local line association under visual degradation.
Recent learning-based designs improve front-end matching, e.g., StereoVO~\cite{kannapiran2023stereo} and GlueStick~\cite{pautrat2023gluestick}, but usually introduce additional computation.

\begin{figure*}
      \centering
	   \includegraphics[width=6.8in]{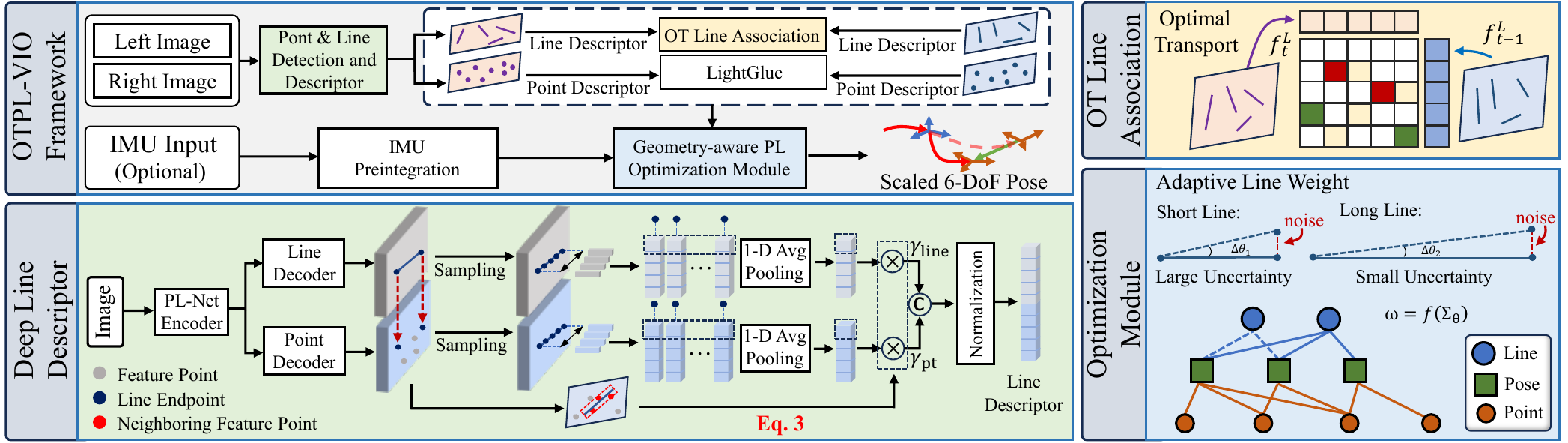}
       \vspace{-0.2cm}
      \caption{Overview and key modules of OTPL-VIO. The upper-left panel shows the stereo point--line VIO pipeline; the remaining panels detail the deep line descriptor, OT-based line association, and reliability-adaptive line weighting.}
      \label{Fig_1}
      \vspace{-0.4cm}
\end{figure*}

\noindent\textbf{Efficient point--line front-ends and the remaining gap.}
AirSLAM~\cite{xu2025airslam} improves efficiency with a unified learning-based point--line front-end and point-guided line association.
While effective when point support is sufficient, point-guided line association can become ambiguous when point features are sparse or unstable, e.g., in low-texture regions and under abrupt illumination changes.

Overall, existing methods reveal a trade-off among robustness, association consistency, and efficiency:
(i) handcrafted line descriptors can degrade under weak gradients,
(ii) heavy learning-based matching can be costly for real-time deployment, and
(iii) point-guided line association may fail when point support collapses.
This motivates explicit lightweight line descriptors, globally normalized line association, and reliability-aware optimization for heterogeneous line quality.

\section{Method}

\subsection{Overview}
As shown in Fig.~\ref{Fig_1}, we present a stereo point--line odometry system with an optional inertial module, combining a learning-based front-end with a factor-graph back-end.
Given a stereo image pair, we use PL-Net~\cite{xu2025airslam} to jointly detect keypoints and line segments.
PL-Net directly outputs the 2D locations of keypoints and line segments, together with per-keypoint descriptors, while it does not provide explicit descriptors for line segments.
Keypoints are matched across frames using LightGlue~\cite{lindenberger2023lightglue} to obtain point correspondences.

Our goal is to strengthen line constraints when point measurements become weak or unstable, particularly in low-texture regions and under abrupt illumination changes.
To this end, we (i) construct lightweight line descriptors by pooling deep features along each segment, (ii) perform OT-based line association with virtual nodes to handle unmatched segments, and (iii) adaptively weight line reprojection factors to reduce the influence of noisy line measurements.
Point, line, and optional IMU factors are jointly optimized in a unified factor graph.

\subsection{Line-Aware Front-end: Descriptor and OT Association}
\label{sec:frontend}

\noindent\textbf{Density-guided hybrid deep line descriptor.}
To improve line discriminability in low-texture regions and under abrupt illumination changes, we build a lightweight segment-level hybrid descriptor by pooling deep features along each detected segment.
We use PL-Net~\cite{xu2025airslam} to extract keypoints and line segments, together with two feature maps $\mathcal{F}_{\text{line}} \in \mathbb{R}^{D_{\text{line}} \times H \times W}$ and $\mathcal{F}_{\text{pt}} \in \mathbb{R}^{D_{\text{pt}} \times H \times W}$,
corresponding to the line-specific and point-specific branches, respectively, where $D_{\mathrm{line}}$ and $D_{\mathrm{pt}}$ are channel dimensions and $H,W$ denote the resolution of the feature map.
Notably, our descriptor requires no additional network or training: it is computed by sampling and pooling PL-Net feature maps, introducing negligible overhead. Although both point- and line-branch feature maps are used, the descriptor is computed along the entire segment rather than induced from discrete point matches.

Given a detected 2D line segment $\mathbf{l}=(\mathbf{p}_s,\mathbf{p}_e)$, we uniformly sample $N_s$ locations $\{s_k\}$ along the segment in feature-map coordinates.
For each branch $b\in\{\text{line, pt}\}$, we obtain per-sample features by bilinear sampling
$\mathbf{v}_k^{(b)}=\text{bilinear}(\mathcal{F}_b, s_k)$.
We then average-pool along the segment and $\ell_2$-normalize:
\begin{equation}
\mathbf{d}_b = \frac{1}{N_s}\sum_{k=0}^{N_s-1}\mathbf{v}_k^{(b)},
\qquad
\mathbf{f}_b = \frac{\mathbf{d}_b}{\|\mathbf{d}_b\|_2}.
\end{equation}
The final descriptor is formed by weighted concatenation of the two branches:
\begin{equation}
\mathbf{d}_\ell =
\left[
\gamma_{\text{line}} \cdot \mathbf{f}_{\text{line}}^\top,\ 
\gamma_{\text{pt}} \cdot \mathbf{f}_{\text{pt}}^\top
\right]^\top,\qquad
\mathbf{f}_\ell = \frac{\mathbf{d}_\ell}{\|\mathbf{d}_\ell\|_2},
\end{equation}
where $\gamma_{\text{line}}$ and $\gamma_{\text{pt}}$ control the relative contributions of the two branches.

Intuitively, segments located in point-rich areas should rely more heavily on the point-specific branch to leverage distinct local structures, whereas those in low-texture regions should depend on the line-specific branch to capture holistic structural context. To achieve this, we introduce a geometry-based weighting scheme driven by local keypoint density. Let $\mathcal{P}$ be the set of all extracted keypoints in the image. For a given line segment $\mathbf{l}$, we count the number of neighboring keypoints $N_{\text{near}}$ whose orthogonal distance to $\mathbf{l}$ is less than a distance threshold $r$. The local point density along the segment is then defined as $\rho = N_{\text{near}} / \|\mathbf{p}_e - \mathbf{p}_s\|_2$. We formulate the adaptive weights as:
\begin{equation}
\gamma_{\text{pt}} = \frac{\rho}{\rho + \rho_0}, \qquad \gamma_{\text{line}} = \frac{\rho_0}{\rho + \rho_0},
\end{equation}
where $\rho_0$ is a baseline density hyperparameter. When sufficient keypoints are observed near a segment, the descriptor is dominated by point-specific features, effectively behaving like point-guided line association \cite{xu2025airslam}. When keypoint support is weak, the descriptor automatically shifts emphasis to the line-specific branch, improving robustness in low-texture regions.

\noindent\textbf{Optimal-transport line association.}
Establishing robust line correspondences is particularly challenging in environments with low or repetitive textures. Under such conditions, local matching methods (e.g., Nearest Neighbor) are susceptible to ambiguities. To address this, we match line segments between two frames via an entropy-regularized OT formulation, which jointly assigns candidate segments under marginal constraints and supports unmatched observations through virtual nodes.
Let $\mathcal{L}^A=\{l^A_i\}_{i=1}^{M}$ and $\mathcal{L}^B=\{l^B_j\}_{j=1}^{N}$ be the line sets in frames $A$ and $B$.
In the OT formulation, we use the projected length of each detected segment in the image as its transport mass, so that the marginal constraints reflect the scale of each line observation:
\begin{equation}
\begin{aligned}
\mathbf{a} &= [a_1,\dots,a_M]^\top, & a_i &= \|\mathbf{p}_{i,s}^A-\mathbf{p}_{i,e}^A\|_2,\\
\mathbf{b} &= [b_1,\dots,b_N]^\top, & b_j &= \|\mathbf{p}_{j,s}^B-\mathbf{p}_{j,e}^B\|_2.
\end{aligned}
\end{equation}
To handle $\sum_i a_i\neq \sum_j b_j$, we augment the marginals with one virtual node on each side:
\begin{equation}
\hat{\mathbf{a}}=[a_1,\dots,a_M,\ \sum_{j=1}^{N}b_j]^\top,\qquad
\hat{\mathbf{b}}=[b_1,\dots,b_N,\ \sum_{i=1}^{M}a_i]^\top.
\end{equation}
Given normalized descriptors computed by Eq.~(2), we define $C_{ij}=1-\langle \mathbf{f}^A_i,\mathbf{f}^B_j\rangle$ for $i\le M,j\le N$ and construct the augmented cost
\begin{equation}
\hat{\mathbf{C}} =
\begin{bmatrix}
\mathbf{C} & \tau\cdot \mathbf{1}_{M} \\
\tau\cdot \mathbf{1}_{N}^\top & 0
\end{bmatrix},
\end{equation}
where $\tau$ is the virtual-node matching cost. We solve
\begin{equation}
\begin{split}
\hat{P}^*=\arg\min_{\hat{P}\ge 0}\ 
&\langle \hat{P},\hat{C}\rangle
+\varepsilon\sum_{i=1}^{M+1}\sum_{j=1}^{N+1}\hat{P}_{ij}(\log \hat{P}_{ij}-1)\\
\text{s.t.}\ 
&\hat{P}\mathbf{1}=\hat{\mathbf{a}},\quad \hat{P}^\top \mathbf{1}=\hat{\mathbf{b}},
\end{split}
\end{equation}
where $\varepsilon$ controls the entropy regularization strength. Let $P$ be the top-left $M\times N$ block of $\hat{P}^*$ and define the row-normalized confidence
\begin{equation}
T_{ij}=\frac{P_{ij}}{a_i+\eta},
\end{equation}
where $\eta$ is a small constant.
We obtain discrete matches by mutual best selection with a confidence threshold $\delta$:
\begin{equation}
j=\arg\max_{k\in[1,N]}T_{ik},\quad
i=\arg\max_{k\in[1,M]}T_{kj},\quad
T_{ij}>\delta.
\end{equation}
The accepted matches form the set $\mathcal{M}_{OT}$. 

\subsection{Geometry-Aware Back-end: Residuals and Reliability-Adaptive Optimization}
\label{sec:backend}

\noindent\textbf{Line representation and reprojection residual.}
We represent a 3D line landmark using Pl\"ucker coordinates $\mathbf{L}=[\mathbf{n}^\top,\mathbf{d}^\top]^\top\in\mathbb{R}^6$ and transform it to the camera frame as $\mathbf L_c=[\mathbf n_c^\top,\mathbf d_c^\top]^\top$ under pose $\mathbf T_{cw}$.
The corresponding 2D line in homogeneous form is
\begin{equation}
\mathbf{l}=[l_1,l_2,l_3]^\top \propto \mathbf{K}^{-\top}\mathbf{n}_c ,
\end{equation}
where $\mathbf{K}$ is the intrinsic matrix.
Given an observed 2D line segment with endpoints $\mathbf{p}_s$ and $\mathbf{p}_e$, let $\tilde{\mathbf{p}}_s$ and $\tilde{\mathbf{p}}_e$ be their homogeneous coordinates. We define the line reprojection residual as endpoint-to-line distances:
\begin{equation}
\mathbf{r}_\ell =
\begin{bmatrix}
\frac{\tilde{\mathbf{p}}_s^\top\mathbf{l}}{\sqrt{l_1^2+l_2^2}},\frac{\tilde{\mathbf{p}}_e^\top\mathbf{l}}{\sqrt{l_1^2+l_2^2}}
\end{bmatrix}^T.
\end{equation}

\noindent\textbf{Reliability-adaptive line weighting.}
Line measurements have heterogeneous reliability across time and scenes.
For each matched pair $(i,j)\in\mathcal{M}_{OT}$, we assign a weight $\omega_{ij}$ that combines geometric stability and track persistence:
\begin{equation}
\omega_{ij}=w_{\mathrm{geo}}\cdot w_{\mathrm{vis}}.
\end{equation}

We model $w_{\mathrm{geo}}$ using the length-dependent orientation uncertainty under isotropic endpoint noise.
Let $\mathbf{d}=\mathbf{p}_e-\mathbf{p}_s$ and $L=\|\mathbf{d}\|_2$.
Assuming independent endpoint noise $\mathcal{N}(0,\sigma_{\mathrm{img}}^2\mathbf{I})$, first-order error propagation yields
\begin{equation}
\sigma_\theta^2 \approx \frac{2\sigma_{\mathrm{img}}^2}{L^2},
\end{equation}
showing that orientation uncertainty decreases with the squared segment length.
To account for intrinsic system noise, we use
\begin{equation}
\sigma_{\theta}^2 = \sigma_{\mathrm{base}}^2 + \frac{\kappa}{L^2},
\end{equation}
where $\kappa$ is a length-scaling parameter reflecting the effective image noise variance. 
We therefore assign a geometric weight inversely proportional to this orientation uncertainty:
\begin{equation}
w_{\mathrm{geo}} = \max\left( \frac{\lambda}{\sigma_{\mathrm{base}}^2 + \kappa / L^2},\ w_{\min} \right),
\end{equation}
where $\lambda$ is a constant scaling factor. For track persistence, we set a visibility weight based on track length:
\begin{equation}
w_{\mathrm{vis}} =
\begin{cases}
1.0, & \text{if } N_{\mathrm{obs}} \ge \tau_{\mathrm{trk}}\\
w_{\min}, & \text{otherwise}.
\end{cases}
\end{equation}
where $N_{\mathrm{obs}}$ is the number of frames in which the segment is successfully tracked. The weighting procedure is summarized in Algorithm~\ref{alg:weight_eval}.

\begin{algorithm}[t]
\caption{Reliability-adaptive line weighting}
\label{alg:weight_eval}
\begin{algorithmic}[1]
\Require Matched pairs $\mathcal{M}_{OT}=\{(i,j)\}$, where $i$ denotes the line track index and $j$ the observation in the current frame; endpoints $(\mathbf{p}_{j,s},\mathbf{p}_{j,e})$ and track length $N_{obs}(i)$
\Ensure Weights $\mathbf{\Omega}=\{\omega_{ij}\}$
\Statex \textbf{Parameters:} $\sigma_{\mathrm{base}}, \kappa, \lambda, w_{\min}, \tau_{trk}$
\State $\mathbf{\Omega}\leftarrow \emptyset$
\For{\textbf{each} $(i,j)\in \mathcal{M}_{OT}$}
    \State $L \leftarrow \|\mathbf{p}_{j,s}-\mathbf{p}_{j,e}\|_2$
    \State $L \leftarrow \max(L, 1.0)$
    \State $\sigma^2_{\theta} \leftarrow \sigma_{\mathrm{base}}^2 + \kappa / L^2$
    \State $w_{\mathrm{geo}} \leftarrow \max\!\left(\lambda / \sigma_{\theta}^2,\ w_{\min}\right)$
    \State $w_{vis} \leftarrow 1.0$
    \If{$N_{obs}(i) < \tau_{trk}$}
        \State $w_{vis} \leftarrow w_{\min}$
    \EndIf
    \State $\omega_{ij} \leftarrow w_{\mathrm{geo}}\cdot w_{vis}$
    \State $\mathbf{\Omega} \leftarrow \mathbf{\Omega}\cup\{\omega_{ij}\}$
\EndFor
\State \Return $\mathbf{\Omega}$
\end{algorithmic}
\end{algorithm}

\noindent\textbf{Factor-graph optimization.}
We perform local bundle adjustment over keyframe poses, 3D points, and 3D lines in a factor graph.
The objective consists of standard point reprojection errors and weighted line reprojection errors (using $\mathbf{r}_\ell$ with weight $\omega_{ij}$), and when inertial measurements are available, we additionally include standard IMU preintegration factors:
\begin{equation}
E = E_{point}+E_{line}+E_{imu},
\end{equation}
where $E_{imu}$ is included only when inertial measurements are available.

\section{Experiment}

\subsection{Implementation Details}
All hyperparameters are fixed across evaluations.
For the density-guided hybrid deep line descriptor, we set the number of samples to $N_s=100$. The adaptive weights are computed using a neighborhood distance threshold of $r=3$ pixels and a baseline density parameter $\rho_0=0.1$. For the OT line association, $\tau = 0.15$, $\varepsilon = 0.05$, $\eta = 10^{-8}$, $\delta = 0.6$.
Here, $\varepsilon$ controls the smoothness of the transport plan, while $\tau$ controls the cost of assigning segments to virtual unmatched nodes.
For reliability-adaptive weighting, we use $\sigma_{\mathrm{base}}=1$, $\lambda=1$, $\kappa=40000$, $\tau_{trk}=3$, and $w_{\min}=0.1$. To quantify feature validity in our analysis, stereo triangulation is deemed successful if the matched point yields a valid horizontal parallax within the physical camera bounds, resulting in a positive depth.

\begin{figure}
    \centering
    \hspace{0.7cm}
    \includegraphics[width=0.8\linewidth]{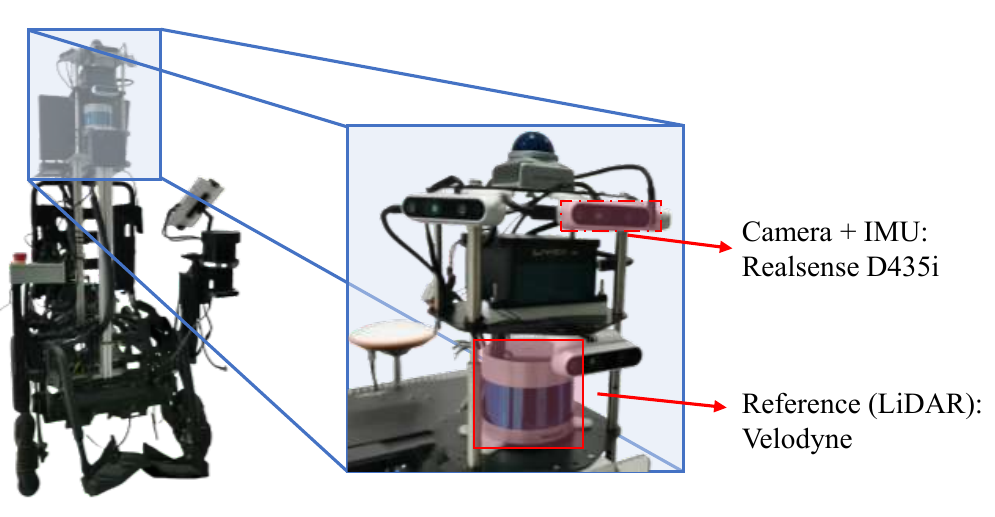}
    \vspace{-0.4cm}
    \caption{Our data-collection platform.}
    \label{Fig_2}
    \vspace{-0.4cm}
\end{figure}

\begin{table*}
\centering
\caption{ATE RMSE [cm] on EuRoC dataset. E, M, and H denote Easy, Medium, and Hard sequences, respectively.
}

\label{tab:euroc_results}
\vspace{-0.15cm}
\resizebox{0.9\linewidth}{!}{
\begin{tabular}{lccccccccccccc}
\toprule
\multirow{3}{*}{Method} & \multicolumn{13}{c}{Sequence} \\
\cmidrule(lr){2-14}
& E & E & M & H & H & E & M & H & E & M & H &  &  \\
& MH01 & MH02 & MH03 & MH04 & MH05 & V101 & V102 & V103 & V201 & V202 & V203 & Avg. \\
\midrule
VINS-Fusion~\cite{qin2018vins}         & 16.30 & 17.80 & 31.60 & 33.10 & 17.50 & 10.20 & 9.90  & 11.20 & 11.00 & 12.40 & 25.20 & 17.84 \\
DROID-SLAM~\cite{teed2021droid}        & 16.30 & 12.10 & 24.20 & 39.90 & 27.00 & 10.30 & 16.50 & 15.80            & 10.20 & 11.50 & 20.40 & 18.56 \\
Struct-VIO~\cite{zou2019structvio}     & 11.90 & 10.00 & 28.30 & 27.50 & 25.60 & 7.50  & 19.70 & 16.10            & 8.10  & 15.20 & 17.70 & 17.05 \\
PLF-VINS~\cite{lee2021plf}             & 14.30 & 17.80 & 22.10 & 24.00 & 26.00 & 6.90  & 9.90  & 16.60            & 8.30  & 12.50 & 18.30 & 16.06 \\
Kimera-VIO~\cite{rosinol2020kimera}    & 11.00 & 10.00 & 16.00 & 24.00 & 35.00 & 5.00  & 8.00  & \underline{7.00}    & 8.00  & 10.00 & 21.00 & 14.09 \\
OKVIS~\cite{leutenegger2015keyframe}   & 19.70 & 10.80 & \underline{12.20} & \underline{13.80} & 27.20 & 4.00  & \textbf{6.70} & 12.00 & 5.50  & 15.00 & 24.00 & 13.72 \\
AirSLAM~\cite{xu2025airslam}           & \textbf{7.40} & \underline{6.00} & \textbf{11.40} & 16.70 & \underline{12.50} & \textbf{3.30} & 13.20 & 23.80 & \underline{3.60} & \underline{8.30} & \underline{16.80} & \underline{11.18} \\
Ours                                   & \underline{8.99} & \textbf{5.09} & 12.56 & \textbf{9.93} & \textbf{10.91} & \underline{3.83} & \underline{6.81} & \textbf{6.94} & \textbf{3.34} & \textbf{8.29} & \textbf{11.97} & \textbf{8.06} \\
\bottomrule
\end{tabular}
}
\vspace{-0.6cm}
\end{table*}

\subsection{Datasets and Evaluation Protocol}
We evaluate the proposed point--line VIO system on public benchmarks and self-collected real-world sequences. The public benchmarks include EuRoC MAV~\cite{burri2016euroc}, UMA-VI~\cite{zuniga2020vi}, and HPatches~\cite{balntas2017hpatches}. EuRoC provides synchronized stereo--inertial sequences in indoor industrial environments with different motion and visual difficulty levels. UMA-VI contains challenging VIO sequences grouped by degradation factors; we focus on the illumination-change and low-texture subsets. HPatches is used only for line association evaluation under illumination and viewpoint changes.

For real-world validation, we collect three indoor sequences using the mobile platform in Fig.~\ref{Fig_2}. Static-1 and Static-2 are static-scene routes with abrupt illumination changes and low-texture regions, while the Dynamic-1 includes pedestrian occlusions and moving line observations. Since external ground truth is unavailable in these real-world scenes, we use an offline loop-closure LIO trajectory as a reference rather than absolute ground truth.

\subsection{Public Benchmarks under Illumination Changes and Low-Texture}
We evaluate on EuRoC using all sequences, and on UMA-VI using the illumination-change subset (10 sequences) and the low-texture subset (5 sequences). For UMA-VI, ground truth is available only at the start and end of each sequence; accordingly, we report results under a no-loop-closure setting and compute the RMSE of the absolute translation error (ATE) only over the available ground-truth segments, rather than over the dense full trajectory.

\begin{figure}[!t]
    \centering
    \includegraphics[width=\linewidth]{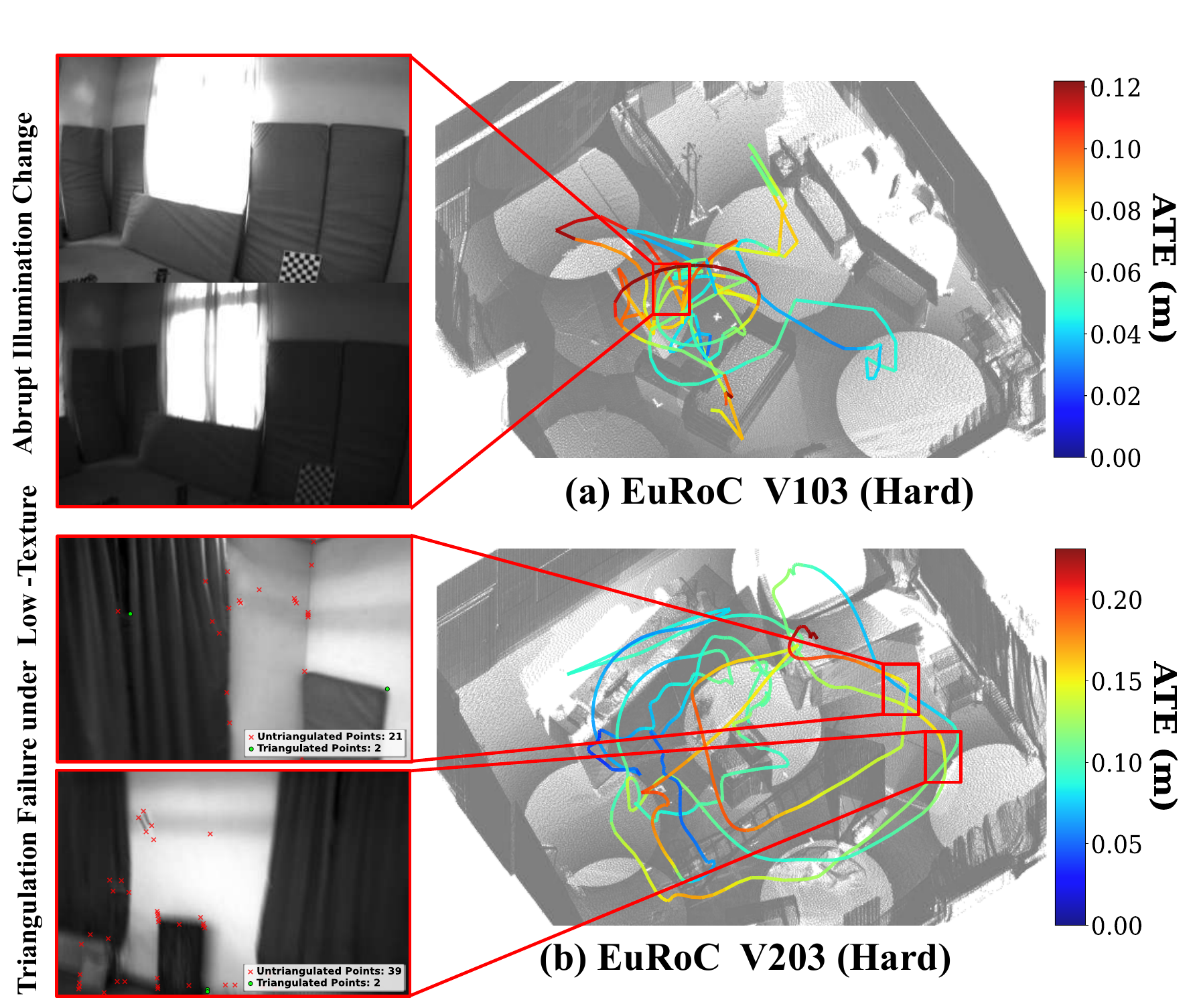}
    \vspace{-0.5cm}
    \caption{Qualitative evaluation on EuRoC V103 and V203. Estimated trajectories are colored by ATE. Red boxes indicate challenging views: abrupt illumination changes in V103 and low-texture regions in V203. Red crosses and green dots denote failed and successful triangulations, respectively.}
    \label{Fig_3}
    \vspace{-0.4cm}
\end{figure}

Table~\ref{tab:euroc_results} reports the EuRoC results. Our method achieves the best overall performance with an average RMSE of 8.06~cm, improving upon the strongest baseline (AirSLAM, 11.18~cm) by 27.9\%. The gain is most evident on the harder sequences (e.g., MH04/MH05, V103 and V203), where point associations become more ambiguous. These improvements stem from our descriptor-aware line representation, OT-based joint line assignment, and reliability-adaptive optimization that stabilizes pose updates when line quality varies. Fig.~\ref{Fig_3} provides qualitative visualizations on EuRoC, illustrating stable tracking across appearance changes and texture-poor segments.

Table~\ref{tab:uma_vi_rmse} reports results on the UMA-VI illumination-change sequences. Our method achieves the best average performance (25.5~cm), reducing RMSE by 42.2\% compared with AirSLAM (44.1~cm). In these scenarios, traditional point-based methods (e.g., ORB-SLAM3) frequently fail due to the instability of ORB features under lighting changes. Similarly, hand-crafted point--line systems (e.g., PL-SLAM) struggle because unstable image gradients degrade LSD extraction and LBD matching. Consequently, the resulting line outliers corrupt the back-end optimization, causing these systems to occasionally underperform point-only baselines. Although AirSLAM employs illumination-robust deep front-ends, its point-guided line association still depends on reliable local point support. Under abrupt lighting changes, point features near structural edges often become less stable, which can disrupt line matching and introduce association outliers. In contrast, our descriptor-based line representation and OT-based matching perform joint assignment over candidate line segments and support unmatched observations under ambiguity and partial observations, while reliability-adaptive optimization suppresses noisy line constraints during pose estimation.

\begin{table}[t]
\centering
\caption{ATE RMSE [cm] on UMA-VI illumination-change sequences.
\NA denotes tracking failure or large drift error.}
\label{tab:uma_vi_rmse}
\vspace{-0.15cm}
\resizebox{\columnwidth}{!}{%
\begin{tabular}{lccccccc}
\toprule
\multirow{2}{*}{Sequence} & PL- & ORB- & Basalt & OKVIS & DROID- & AirSLAM & \multirow{2}{*}{Ours} \\
& SLAM \cite{gomez2019pl} & SLAM3 \cite{campos2021orb} & \cite{usenko2019visual} & \cite{leutenegger2015keyframe} & SLAM \cite{teed2021droid} & \cite{xu2025airslam} & \\
\midrule
conf-csc1   & 269.7 & \NA  & 127.0 & 111.8 & 71.1  & \underline{49.0} & \textbf{37.3} \\
conf-csc2   & 159.6 & \NA  & 68.2  & 47.0  & 13.5  & \textbf{9.1}     & \underline{11.6} \\
conf-csc3   & \NA   & 42.6 & 46.9  & \underline{8.8} & 72.4 & \underline{8.8} & \textbf{3.7} \\
lab-rev     & \NA   & \textbf{6.3} & 48.6 & 86.1 & \underline{36.4} & 50.4 & 38.7 \\
lab-csc     & \NA   & \NA  & 40.3  & 57.9  & \underline{31.9} & 97.9 & \textbf{25.9} \\
long-eng    & \NA   & \NA  & 504.6 & 300.5 & \NA   & \underline{180.1} & \textbf{104.9} \\
third-csc1  & 447.8 & 86.3 & 42.0  & 28.7  & \underline{4.8}  & 7.0  & \textbf{3.5} \\
third-csc2  & 606.8 & 14.9 & 59.0  & 27.1  & 89.0  & \underline{12.7} & \textbf{11.8} \\
two-csc1    & \NA   & \NA  & 76.0  & 15.4  & 34.1  & \underline{6.6}  & \textbf{4.2} \\
two-csc2    & \NA   & \NA  & 121.1 & 67.9  & 29.9  & \underline{19.0} & \textbf{13.1} \\
\midrule
Avg.         & \NA & \NA & 113.4 & 75.1 & \NA & \underline{44.1} & \textbf{25.5} \\
\bottomrule
\end{tabular}%
}
\vspace{-0.3cm}
\end{table}

\begin{figure}[t]
    \centering
    \includegraphics[width=0.9\linewidth]{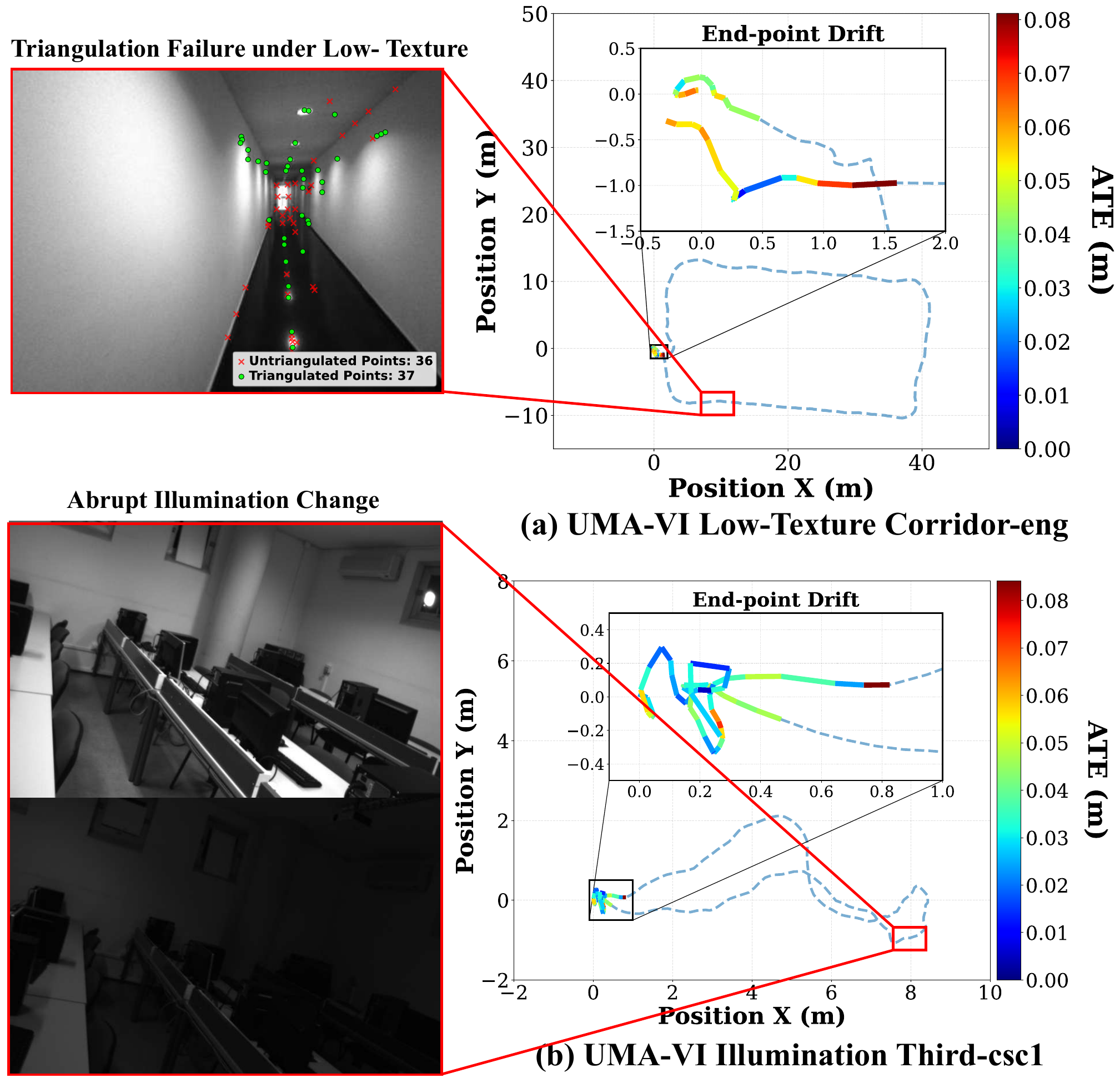}
    \vspace{-0.2cm}
    \caption{Qualitative results on UMA-VI. Estimated trajectories are colored by ATE where reference poses are available; dashed lines indicate segments without reference poses. Insets show accumulated end-point drift. Red boxes mark low-texture and illumination-changing views.}
    \label{Fig_4}
    \vspace{-0.2cm}
\end{figure}

Table~\ref{tab:uma_lowtexture} summarizes results on the UMA-VI low-texture sequences. These sequences are particularly challenging because weak textures lead to under-constrained geometry and unreliable point support.
In addition, the UMA-VI low-texture subset contains repetitive indoor patterns, making it a practical test case for repeated-pattern ambiguity.
For methods that rely on point-guided line association, unreliable point support can further propagate to line matching, weakening the resulting line constraints. Although the deep front-end can still detect dense keypoints in such regions, the lack of distinctive texture often prevents stable stereo matching, leading to unreliable triangulation and weak geometric constraints (Fig.~\ref{Fig_4}). By introducing explicit line descriptors, robust OT-based line association, and reliability-adaptive weighting, our method achieves markedly improved accuracy and robustness over representative point--line baselines.

\begin{figure}[t]
    \centering
    \includegraphics[width=0.7\linewidth]{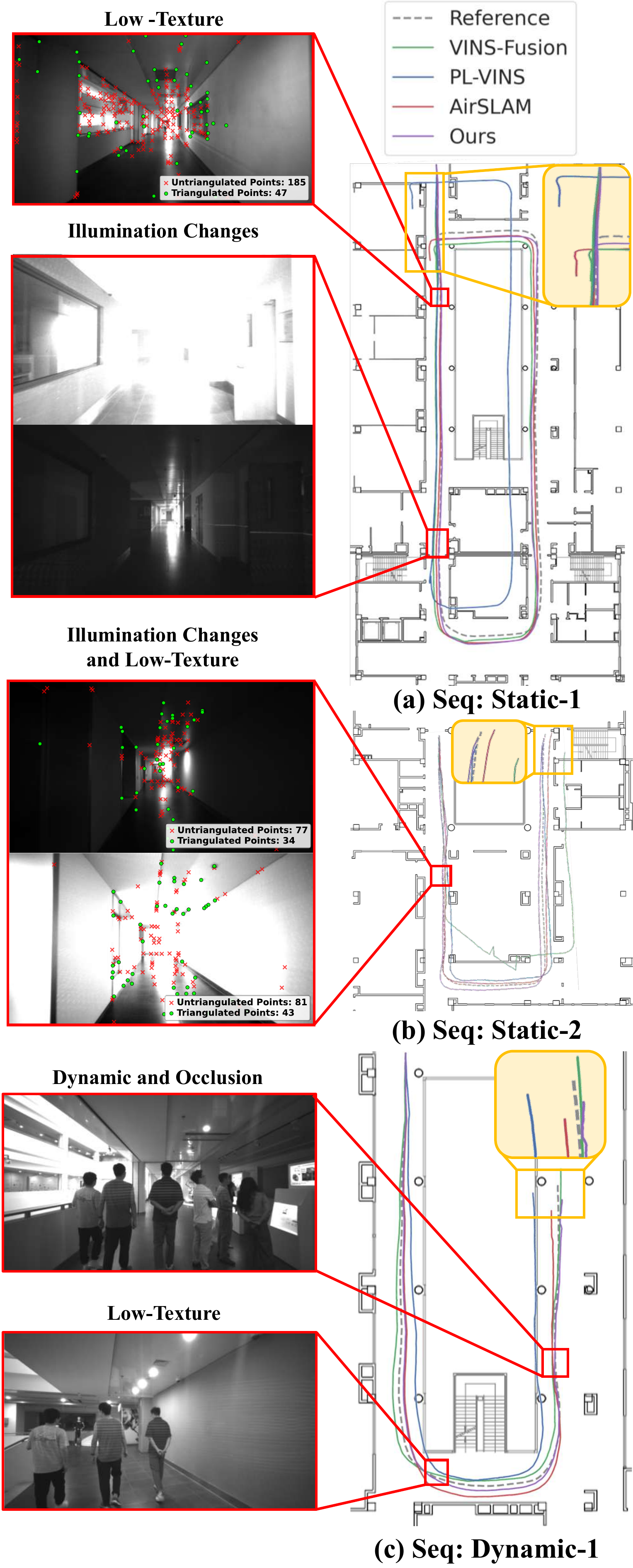}
    \vspace{-0.2cm}
    \caption{Qualitative results on real-world sequences. Estimated trajectories are overlaid on the floor plans for Static-1, Static-2, and Dynamic-1. Red boxes show representative low-texture, illumination-changing, dynamic, and occlusion scenes; orange boxes enlarge trajectory endpoints.}
    \label{Fig_5}
    \vspace{-0.2cm}
\end{figure}

\begin{table}
\caption{ATE RMSE [cm] on UMA-VI Low-Texture sequences. 
}
\label{tab:uma_lowtexture}
\vspace{-0.15cm}
\centering
\resizebox{\linewidth}{!}{%
\begin{tabular}{lcccccc}
\toprule
Sequence & VINS-Fusion\cite{qin2018vins} & PL-VINS\cite{fu2020plvins} & AirSLAM\cite{xu2025airslam} & Ours \\
\midrule
corridor-eng & 39.92 & 34.52 & \underline{6.20} & \textbf{5.12} \\
parking-csc1 & \underline{33.14} & 55.58 & 48.99 & \textbf{20.19} \\
parking-csc2 & \underline{26.03} & 50.24 & 38.82 & \textbf{18.05} \\
class-csc1 & 19.23 & 22.03 & \underline{18.70} & \textbf{10.29} \\
class-csc2 & 23.23 & 29.81 & \underline{17.50} & \textbf{4.37} \\
\midrule
Avg. & 28.31 & 38.43 & \underline{26.04} & \textbf{11.60} \\
\bottomrule
\end{tabular}%
}
\vspace{-0.4cm}
\end{table}

\begin{table}[t]
\caption{ATE RMSE [cm] on real-world sequences. Illum. and occl. denote illumination and occlusion, respectively.
}
\label{tab:real-world}
\vspace{-0.3cm}
\centering
\resizebox{\columnwidth}{!}{%
\begin{tabular}{llcccc}
\toprule
Sequence & Scenario & VINS-Fusion & PL-VINS & AirSLAM & Ours \\
\midrule
Static-1 & Low texture + illum. change & \underline{56.65} & 299.82 & 60.96 & \textbf{39.99} \\
Static-2 & Low texture + illum. change & 273.12 & 90.08 & \underline{37.77} & \textbf{25.71} \\
Dynamic-1 & Low texture + dynamic occl. & \underline{43.50} & 116.92 & 52.61 & \textbf{37.37} \\
\bottomrule
\end{tabular}%
}
\vspace{-0.3cm}
\end{table}

\begin{table}[t]
\centering
\caption{Ablation study on UMA-VI subsets [cm]. DLD: deep line descriptor; OTA: OT-based association; RAW: reliability-adaptive weighting.
}
\label{tab:ablation}
\vspace{-0.15cm}
\resizebox{\columnwidth}{!}{%
\begin{tabular}{lccc|ccc}
\toprule
Method & DLD & OTA & RAW & Low-Tex. & Illum. & Avg. \\
\midrule
\multicolumn{7}{l}{\textbf{Baseline}} \\
\hspace{1em}AirSLAM \emph{(point-guided line association)} &  &  &  & 26.04 & 44.06 & 38.05 \\
\addlinespace[1pt]
\multicolumn{7}{l}{\textbf{Ablation variants}} \\
\hspace{1em}LBD + NN &  &  &  & 18.57 & 31.36 & 27.09 \\
\hspace{1em}LBD + NN + RAW &  &  & \checkmark & 16.77 & 30.73 & 26.07 \\
\hspace{1em}DLD + NN & \checkmark &  &  & 17.87 & 30.82 & 26.50 \\
\hspace{1em}DLD + NN + RAW & \checkmark &  & \checkmark & 11.89 & 30.03 & 23.98 \\
\hspace{1em}DLD + OTA & \checkmark & \checkmark &  & 15.53 & 29.33 & 24.73 \\
\hspace{1em}\textbf{DLD + OTA + RAW (Full)} & \checkmark & \checkmark & \checkmark & \textbf{11.60} & \textbf{25.47} & \textbf{20.85} \\
\bottomrule
\end{tabular}%
}
\vspace{-0.4cm}
\end{table}

\subsection{Real-World Deployment}
We further validate the proposed system on real-world indoor sequences collected by our mobile platform (Fig.~\ref{Fig_2}). 
The evaluation includes two static sequences and one dynamic sequence. Static-1 and Static-2 contain low-texture regions and abrupt illumination changes, while Dynamic-1 contains low-texture regions with pedestrian occlusions and moving line observations. We report ATE-RMSE with respect to the reference trajectory.

Table~\ref{tab:real-world} summarizes the results. Our method achieves the lowest RMSE on all three sequences. 
The results on Static-1 and Static-2 show improved robustness under low texture and illumination changes, while Dynamic-1 further evaluates the influence of moving line observations and partial occlusions. Although the system does not explicitly segment dynamic objects, transient dynamic line observations are indirectly suppressed through association consistency, track persistence, reliability-adaptive weighting, and robust back-end optimization. Nevertheless, scenes dominated by large moving objects or long stable dynamic edges remain challenging.
Fig.~\ref{Fig_5} shows qualitative results on the real-world sequences, where the trajectories remain stable through illumination transitions, extended low-texture segments, and pedestrian occlusions.


\subsection{Ablation Study}
We conduct ablation studies to isolate the effects of the line descriptor, line association strategy, and reliability-adaptive optimization, while keeping the point front-end fixed as PL-Net keypoints with LightGlue matching. For line descriptors, we compare handcrafted \textbf{LBD} with the proposed deep line descriptor \textbf{DLD}; LBD uses Hamming distance, while DLD uses cosine similarity. For line association, we compare nearest-neighbor matching \textbf{NN} with the proposed OT-based association \textbf{OTA}. We denote reliability-adaptive weighting as \textbf{RAW}.
We evaluate AirSLAM as the point-guided line association baseline. For descriptor-based variants, we compare LBD+NN and DLD+NN, as well as their RAW-enhanced versions, LBD+NN+RAW and DLD+NN+RAW. We further evaluate DLD+OTA without RAW and the full system, DLD+OTA+RAW.


\begin{table}[t]
\centering
\caption{Line association results on HPatches.}
\label{tab:line_matching}
\vspace{-0.15cm}
\scriptsize
\setlength{\tabcolsep}{3.2pt}
\renewcommand{\arraystretch}{1.05}
\begin{tabular}{l ccc ccc ccc}
\toprule
\multirow{2}{*}{Method}
& \multicolumn{3}{c}{Illumination Changes}
& \multicolumn{3}{c}{Viewpoint Changes}
& \multicolumn{3}{c}{Overall} \\
\cmidrule(lr){2-4}
\cmidrule(lr){5-7}
\cmidrule(lr){8-10}
& P$\uparrow$ & R$\uparrow$ & F$\uparrow$
& P$\uparrow$ & R$\uparrow$ & F$\uparrow$
& P$\uparrow$ & R$\uparrow$ & F$\uparrow$ \\
\midrule
LBD+NN
& 38.44 & 78.02 & 51.51
& 11.08 & 54.89 & 18.44
& 25.93 & 72.08 & 38.14 \\
DLD+NN
& 44.41 & \textbf{90.13} & 59.50
& 17.15 & \textbf{84.96} & 28.55
& 31.94 & \textbf{88.81} & 46.98 \\
DLD+OTA
& \textbf{55.18} & 67.63 & \textbf{60.77}
& \textbf{20.99} & 46.71 & \textbf{28.97}
& \textbf{42.00} & 62.26 & \textbf{50.16} \\
\bottomrule
\end{tabular}
\vspace{-0.1cm}
\end{table}

\begin{figure}[t]
    \centering
    \includegraphics[width=\linewidth]{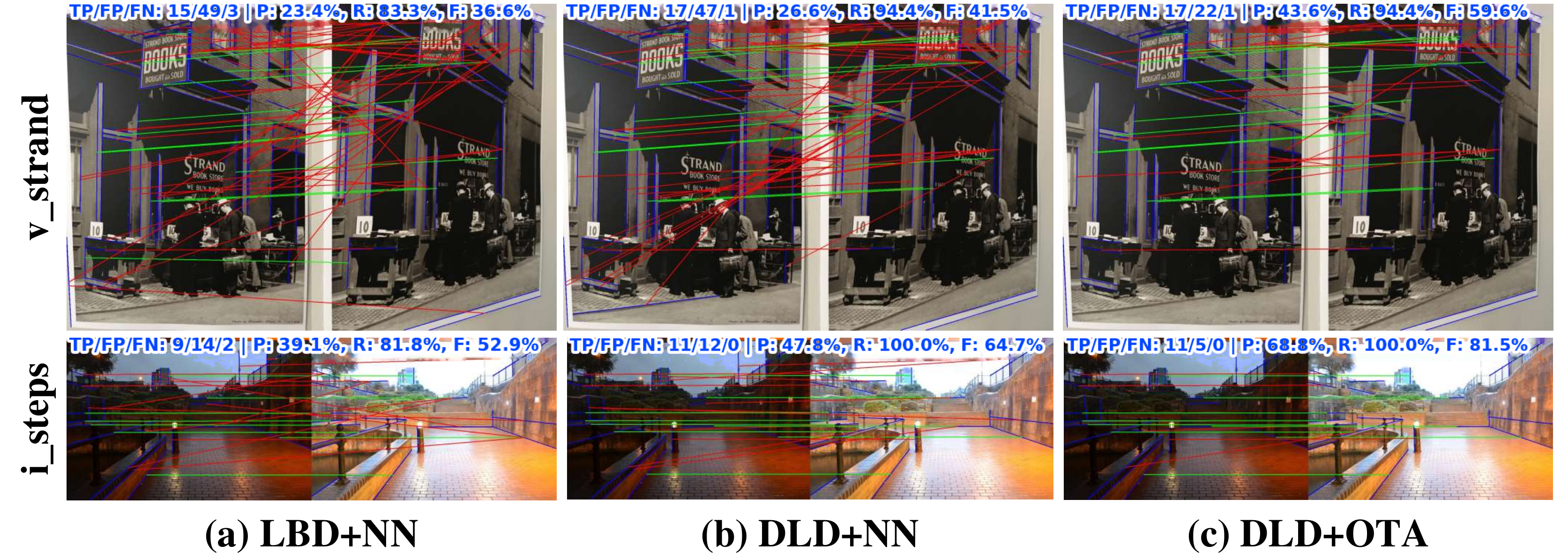}
    \vspace{-0.7cm}
    \caption{Qualitative line association examples on HPatches. The two rows show viewpoint- and illumination-change examples, respectively. Blue segments denote detected lines, while green and red connections indicate true positives and false positives, respectively.}
    \label{Fig_6}
    \vspace{-0.4cm}
\end{figure}

\begin{figure}[t]
    \centering
    \includegraphics[width=\linewidth]{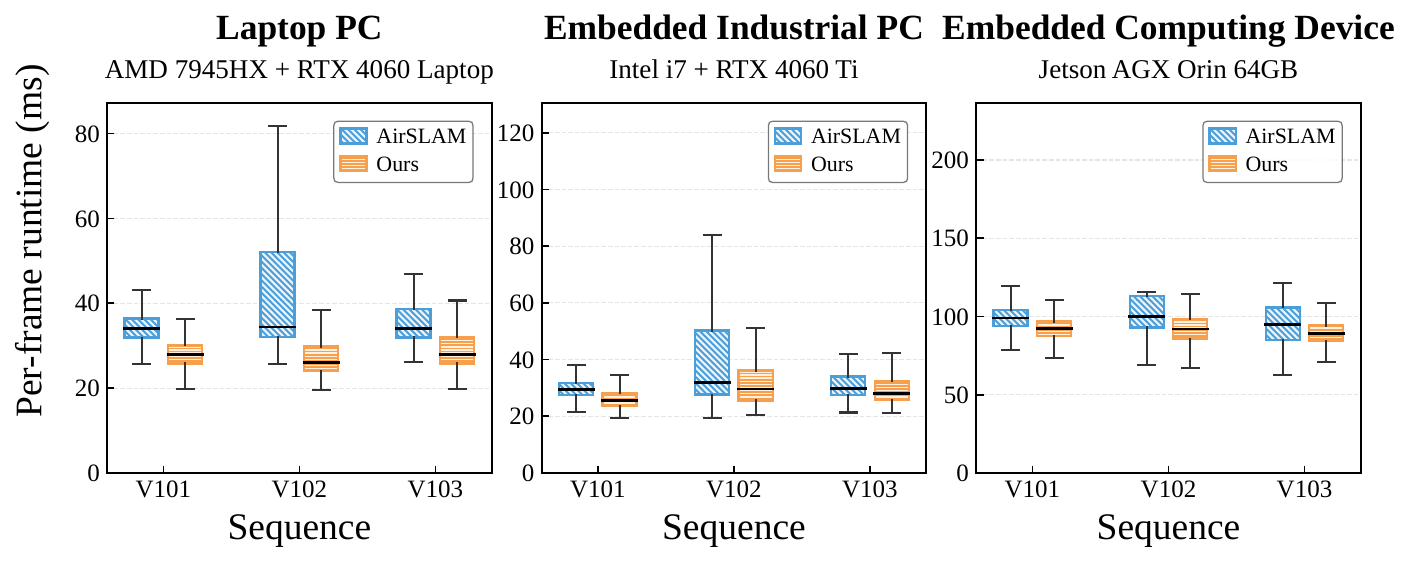}
    \vspace{-0.7cm}
    \caption{Cross-platform runtime on EuRoC V1 sequences. Boxplots show per-frame runtimes of AirSLAM and our method on V101, V102, and V103.}
    \label{Fig_7}
    \vspace{-0.1cm}
\end{figure}

\begin{table}[t]
\centering
\caption{Average ATE RMSE [cm] on EuRoC with loop closure.
}
\label{tab:euroc_lc}
\vspace{-0.15cm}
\resizebox{\columnwidth}{!}{%
\begin{tabular}{lcccccccc}
\toprule
\multirow{2}{*}{Seq.} & Maplab & PL- & ORB- & Basalt & DVI- & DROID- & AirSLAM & \multirow{2}{*}{Ours} \\ & \cite{cramariuc2022maplab}
& SLAM \cite{gomez2019pl} & SLAM3 \cite{campos2021orb} & \cite{usenko2019visual} & SLAM \cite{peng2024dvi} & SLAM \cite{teed2021droid} & \cite{xu2025airslam} & \\
\midrule
Avg.         & 7.1 & 6.1 & 3.5 & 5.2 & 5.1 & \textbf{2.4} & 3.0 & \underline{2.8} \\
\bottomrule
\end{tabular}%
}
\vspace{-0.1cm}
\end{table}

\begin{table}[!t]
\centering
\caption{Average ATE RMSE [cm] on EuRoC and UMA-VI low-texture subset without IMU. 
}
\label{tab:wo_imu}
\vspace{-0.15cm}
\resizebox{0.45\linewidth}{!}{%
\begin{tabular}{lcc}
\toprule
Method & EuRoC & UMA-VI \\
\midrule
AirSLAM \cite{xu2025airslam} & 23.57 & 47.28 \\
Ours    & \textbf{22.95} & \textbf{38.24} \\ 
\bottomrule
\end{tabular}
}
\vspace{-0.5cm}
\end{table}

Table~\ref{tab:ablation} summarizes the ablation results on the UMA-VI low-texture and illumination-change subsets. The full system achieves the best performance on both subsets and the lowest sequence-level average error. Compared with AirSLAM, descriptor-based line association improves the results on both subsets, showing the benefit of using line constraints independently when point support becomes weak or unstable. The ablation variants show that DLD improves line representation, OTA improves front-end line association, and RAW improves back-end robustness by down-weighting unreliable line constraints. The two subsets further show different degradation characteristics. On the low-texture subset, RAW already suppresses many short or fragmented line measurements, so the additional gain from OTA after RAW is relatively small. In contrast, illumination changes make cross-frame association more ambiguous; therefore, combining OTA with RAW gives a clearer improvement. These results show that OTA and RAW are complementary rather than redundant: OTA reduces association ambiguity in the front-end, while RAW controls the back-end influence of the remaining unreliable line constraints.




\subsection{Line Association Evaluation}
We further evaluate the proposed line association module on HPatches~\cite{balntas2017hpatches}, which includes illumination and viewpoint changes. Line segments are independently detected by PL-Net, and the provided homography is used only to generate pseudo ground-truth correspondences. A predicted match is considered correct when the projected reference line overlaps with the target line and satisfies the endpoint reprojection error threshold of 4 pixels and the angle threshold of $2^\circ$. We report Precision (P), Recall (R), and F-score (F) for LBD+NN, DLD+NN, and DLD+OTA.

As shown in Table~\ref{tab:line_matching}, DLD+NN outperforms LBD+NN, confirming the benefit of the proposed descriptor. With the same DLD descriptor, DLD+OTA further improves precision and F-score on both subsets and overall. Although its recall is lower, OTA with virtual nodes can reject ambiguous or unmatched segments instead of forcing correspondences. This provides fewer but more reliable matches, which is more suitable for pose estimation, where false correspondences can introduce inconsistent back-end constraints. Qualitative examples are shown in Fig.~\ref{Fig_6}.


\subsection{Run-time Analysis}
To evaluate computational efficiency and deployment feasibility, we conduct a cross-platform runtime analysis on the EuRoC V1 sequences. We compare AirSLAM and our method on one laptop PC and two industrial/embedded computing platforms. Loop closure is disabled for both methods to ensure a fair odometry-level comparison.

Fig.~\ref{Fig_7} reports the per-frame runtime distributions on V101, V102, and V103. Across the three platforms, our method achieves comparable or lower runtime than AirSLAM and shows a stable runtime profile.  Although runtime increases on the Jetson AGX Orin, our method still runs at around 10 Hz on this platform, indicating practical real-time capability on embedded robotic hardware.

\subsection{Additional Analyses}

\noindent\textbf{Effect of loop closure on EuRoC.}
In addition to the odometry setting used in the main experiments, we evaluate a loop-closure-enabled variant on EuRoC. To ensure a fair comparison, we integrate the same loop-closure module as AirSLAM~\cite{xu2025airslam} into our system.
Loop closure is not a contribution of this work; our proposed modules are used in the odometry front-end and local back-end.
Table~\ref{tab:euroc_lc} shows that loop closure substantially enhances long-term consistency and reduces accumulated drift. With loop closure enabled, our system achieves competitive accuracy against SOTA SLAM baselines under the same loop-closure setting.

\noindent\textbf{Stereo odometry without IMU.}
We also evaluate a vision-only variant by disabling IMU factors and running stereo point--line odometry.
This setting isolates the contribution of line association and reliability-aware optimization when inertial constraints are unavailable.
Results on EuRoC and UMA-VI low-texture subset are summarized in Table~\ref{tab:wo_imu}, indicating that the proposed line descriptor and OT association remain effective, while IMU availability provides additional stabilization in aggressive motions.

\section{Conclusion}
This paper presented OTPL-VIO, a robust and efficient stereo point--line odometry system designed for challenging indoor scenarios where low-texture regions and abrupt illumination changes frequently co-exist. Our approach equips line segments with a deep line descriptor, performs OT-based joint line association via an entropy-regularized optimal transport formulation, and stabilizes back-end estimation through reliability-adaptive weighting. Extensive evaluations on public benchmarks and real-world deployments demonstrate improved accuracy and robustness over representative baselines while maintaining real-time performance. The current system is mainly targeted at structured scenes with persistent line observations; in line-sparse or highly unstructured environments, the benefit of line constraints may decrease. Future work will explore tighter temporal consistency in line association, monocular extension, and larger-scale long-term mapping with loop closure.

\bibliographystyle{IEEEtran}
\bibliography{main}

\newpage

\vfill

\end{document}